\documentclass{llncs}
\usepackage{booktabs,caption,fixltx2e}
\usepackage[flushleft]{threeparttable}
\usepackage{graphicx}
\usepackage{makeidx} 
\usepackage{amsmath}
\usepackage{multirow}

\usepackage{amsfonts}%
\usepackage{amssymb}%
\usepackage[english,oldcommands,plain,lined]{algorithm2e}
\usepackage{graphics}
\usepackage{subfig}
\usepackage[labelfont=bf]{caption}
\usepackage[font={small}]{caption}
\usepackage{transparent}
\usepackage{color}
\usepackage{breqn}
\usepackage{epstopdf}
\usepackage{wrapfig}

\usepackage{color, colortbl}
\definecolor{LightCyan}{rgb}{0.88,1,1}
 \usepackage{cite}
\usepackage[font=small,labelfont=bf]{caption}
\usepackage{enumitem}

\begin{document}

\title{Deep Residual Hashing} 

%
\author{Sailesh Conjeti\inst{1} \and Abhijit Guha Roy\inst{1,2} \and Amin Katouzian\inst{3} \and Nassir Navab\inst {1,4}}
\authorrunning{S. Conjeti et al.} 
\institute{
Computer Aided Medical Procedures, Technische Universit\"{a}t M\"{u}nchen, Germany.\\
\and
Indian Institute of Technology, Kharagpur, WB India.\\
\and
IBM Almaden Research Center, Almaden, USA.\\
\and
Computer Aided Medical Procedures, Johns Hopkins University, USA.}

\maketitle 
\begin{abstract}
Hashing aims at generating highly compact similarity preserving code words which are well suited for large-scale image retrieval tasks. 
 Most existing hashing methods first encode the images as a vector of hand-crafted features followed by a separate binarization step to generate hash codes. This two-stage process may produce sub-optimal encoding. In this paper, for the first time, we propose a deep architecture for supervised hashing through residual learning, termed Deep Residual Hashing (DRH), for an end-to-end simultaneous representation learning and hash coding. The DRH model constitutes four key elements: (1) a sub-network with multiple stacked residual blocks; (2) hashing layer for binarization; (3) supervised retrieval loss function based on neighbourhood component analysis for similarity preserving embedding; and (4) hashing related losses and regularisation to control the quantization error and improve the quality of hash coding. We present results of extensive experiments on a large public chest x-ray image database with co-morbidities and discuss the outcome showing substantial improvements over the latest state-of-the art methods.

\end{abstract}

\section{Introduction}
\label{sec:intro}

Content-based image retrieval (CBIR) aims at effectively indexing and mining large image databases such that given an unseen query image we can effectively retrieve images that are similar in content. 
With the deluge in medical imaging data, there is a need to develop CBIR systems that are both fast and efficient.
 However, in practice, it is often infeasible to exhaustively compute similarity scores between the query image and each image within the database. Adding to the challenge of scalability of CBIR systems is the less understood semantic gap between the visual content of the image and the associated expert annotations~\cite{lai2015}. To address these challenges, hashing based CBIR systems have come to a forefront where the system indexes each image with a compact similarity preserving binary code that could be potentially leveraged for very fast retrieval.

 Towards this end, we propose an end-to-end one-stage deep residual hashing (DRH) network to directly generate hash codes from input images. Specifically, the DRH model constitutes of a sub-network with multiple residual convolutional blocks for learning discriminative image representations followed by a fully-connected hashing layer to generate compact binary embeddings. Through extensive validation, we demonstrate that DRH learns discriminative hash codes in an end-to-end fashion and demonstrates high retrieval quality on standard chest x-ray image databases.
The existing hashing methods proposed for efficient encoding and searching approaches have been proposed for large scale retrieval in machine learning and medical image computing can be categorised into: \textbf{(1)} shallow learning based hashing methods like Locality Sensitive Hashing (LSH)~\cite{lsh}), data-driven methods \textit{e.g.} Iterative Quantization (ITQ)~\cite{itq}, Kernel Sensitive Hashing~\cite{ksh}, Circulent Binary Embedding (CBE)~\cite{cbe}, Metric Hashing Forests (MHF)~\cite{mhf}; 
 \textbf{(2)} hashing using deep architectures (only binarization without feature learning) including Restricted Boltzmann Machines in semantic hashing~\cite{seh}, autoencoders in supervised deep hashing~\cite{liong2015} \textit{etc.} and \textbf{(3)} application-specific hashing methods including weighted hashing for histopathological image search~\cite{zhang2015}, binary code tagging for chest X-ray images~\cite{binaryTag}, forest based hashing for neuron images~\cite{mesbah2015}, to name a few.


\begin{figure}[!t]
\centering
\includegraphics[width =0.9\textwidth]{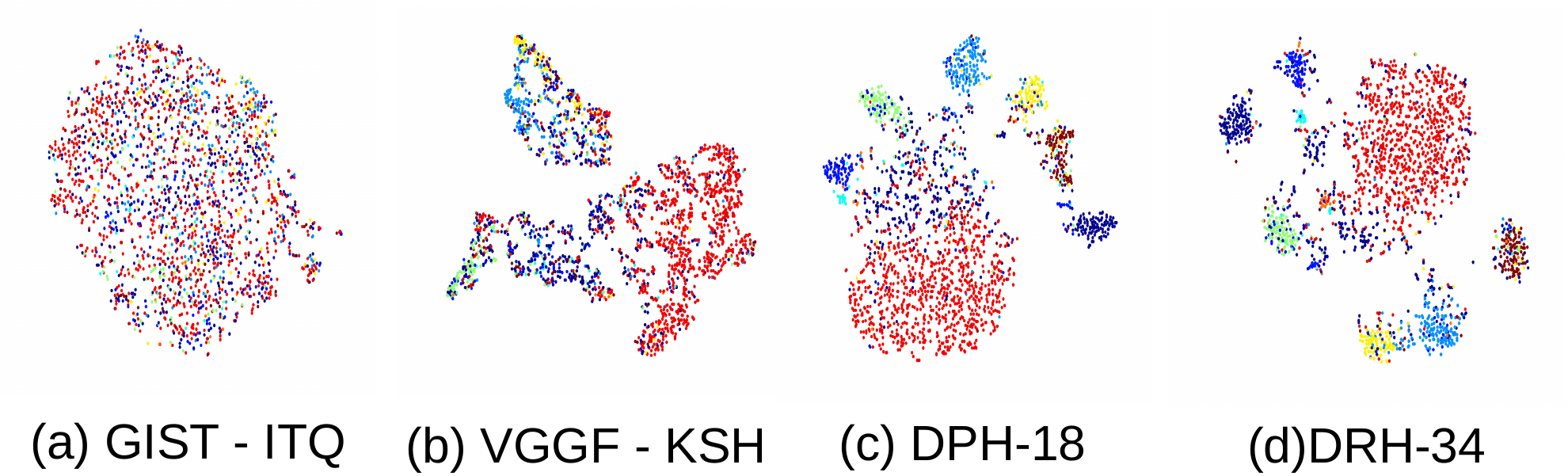}
\caption{tSNE embeddings of the hash codes generated by the proposed and comparative methods. Color indicates different classes. The figure needs to be viewed in color.}

\label{fig:netTSNE}
\end{figure}

\section{Motivation and Contributions}
\label{sec:motivation}

The ultimate objective of earning similarity preserving hashing functions is to generate embeddings in a latent Hamming space such that the class-separability is preserved while embedding and local neighborhoods are well defined and semantically relevant. This can be visualized in 2D by generating the t - Stochastic Neighborhood Embedding (t-SNE)~\cite{tSNE} of unseen test data post learning like shown in Fig.~\ref{fig:netTSNE}. Starting from Fig..~\ref{fig:netTSNE}(a) which is generated by a purely un-superivsed setting we aim at moving towards Fig..~\ref{fig:netTSNE}(d) which is closer to an ideal embedding. In fact, Fig.~\ref{fig:netTSNE} represents the results of our proposed DRH approach in comparison to other methods and baselines. 
%

\noindent
\textbf{Hand-crafted features}: Conventional hashing methods including LSH, ITQ, KSH, MHF \textit{etc.} perform encoding in two stages: firstly, generating a vector of hand-crafted descriptors and a second stage involving hashing learning to preserve the captured semantics in a latent Hamming space. These two independent stages may lead to sub-optimal results as the image descriptors may not be tailored for hashing. Moreover, hand-crafting requires significant domain knowledge and extensive parameter tuning which is particularly undesirable.

\noindent
\textbf{Conventional deep learning}:  Using point-wise loss-functions like cross-entropy, hinge loss \textit{etc.} for training (/ finetuning) deep networks may not lead to feature representations that are sufficiently optimal for the task of retrieval as they do not consider crucial pairwise relationships between instances~\cite{zhu2016}.

\noindent
\textbf{Simultaneous feature learning and hashing}: Recently, with the advent of deep learning for hashing we are able to perform effective end-to-end learning of binary representations directly from input images. These include deep hashing for compact binary code learning~\cite{liong2015}, deep hashing network for effective similarity retrieval~\cite{zhu2016}, simultaneous feature learning and hashing~\cite{lai2015} \textit{etc.} to name a few. However, a crucial disadvantage of these deep learning for hashing methods is that with very deep versions of these networks accuracy gets saturated and often degrades~\cite{ResNet}. In addition to this, the continuous relaxation of hash codes to train deep networks to be able to learn with more viable continuous optimisation methods (gradient-descent based methods) could potentially lead to uncontrolled quantization and distance approximation errors during binarization. 

In an attempt to redress the above short-comings of the existing approaches, we make the following contributions with our work: \textbf{1)} We, for the first lime, design a novel deep hash function learning framework using deep residual networks for representation learning; \textbf{2)} We introduced a neighborhood component analysis-inspired loss suitably tailored for learning discriminative hash codes; \textbf{3)} We leverage multiple hashing related losses and regularizations to control the quantization error while binarization of hash codes and to encourage hash codes to be maximally independent of each other; and \textbf{4)} Clinically, to the best of our knowledge, this is the first retrieval work on medical images (specifically, chest x-ray images) to discuss co-morbidities \textit{i.e.} co-occuring manifestations of multiple diseases. 
The paper also aims at encouraging further discussion on the following aspects of CBIR through DRH:
\begin{enumerate}[leftmargin=*]
\item \textbf{Trainability}: How do we train very deep neural networks for hashing? Does introducing residual connections aid in this process?
\item \textbf{Representability}: Do networks tailored for the dataset at hand learn better representations over transfer learning ? 
\item \textbf{Compactness}: Do highly compact binary representations effectively compress the desired semantic content within an image? Do loss functions to control quantization error while binarzing aid in improved hash coding?
\item \textbf{Semantic-similarity preservation}: Do we learn hash codes such that neighbourhoods in the Hamming space comprise of semantically similar instances?
\item \textbf{Joint Optimisation}: Does end-to-end implicit learning of hash codes work better than a two stage learning process where the images are embedded to a latent space and then quantized explicitly \textit{via} hashing?
\end{enumerate}


\begin{figure}[t]
\centering
\includegraphics[width =0.95\textwidth]{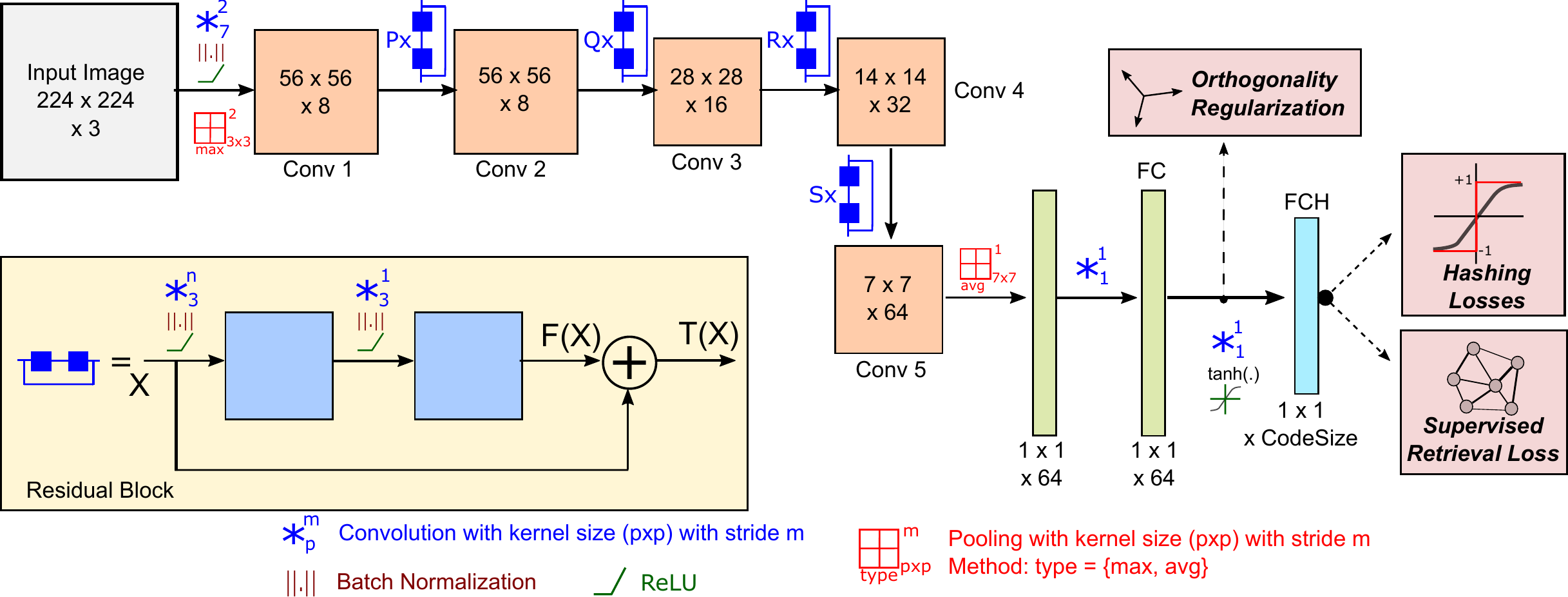}

\caption{Network architecture for deep residual hashing (DRH) with a hash layer. For a 18 - layer network, the number stacked residual blocks are: P = 2, Q = 2, R = 2 and S  = 2. Likewise, for a 34 - layer network, P = 3, Q  = 4, R = 6 and S = 3. The inset image on the left corner is a schematic illustration of a residual block.}
\label{fig:netArchDRH}
\end{figure}

\section{Methodology}


An ideal hashing method should generate codes that are compact, similarity preserving and easy to compute representations (typically, binary in nature), which can be leveraged for accurate search and fast retrieval~\cite{lsh}. The desired similarity preserving aspect of the hashing function implies that \textit{semantically similar images are encoded with similar hash codes}. Mathematically, hashing aims at learning a mapping $\mathcal{H}: \mathcal{I} \rightarrow \left \{ -1, 1 \right \}^{K}$, such that an input image $\mathcal{I}$ can be encoded into a $K$ bit binary code $\mathcal{H}(\mathcal{I})$. In hashing for image retrieval, we typically define a similarity matrix $\mathcal{S} = \left \{ s_{ij} \right \}$, where $s_{ij} = 1$ implies images $\mathcal{I}_{i}$ and $\mathcal{I}_{j}$ are similar and $s_{ij} = 0$ indicates they are dissimilar. Similarity preserving hashing aims at learning an encoding function $\mathcal{H}$ such that the similarity matrix $\mathcal{S}$ is maximally-preserved in the binary hamming space.

\subsection{Architecture for deep residual hashing}

We start with a deep convolutional neural network architecture inspired in part by the seminal ResNet architecture proposed for image classification by He \textit{et al.}~\cite{ResNet}. As shown in Fig.~\ref{fig:netArchDRH}, the proposed architecture consists of the a convolutional layer (Conv 1) followed by a sequence of residual blocks (Conv 2-5) and terminating in a final fully connected hashing (FCH) layer for hash code-generation. The unique advantages offered by the proposed ResNet architecture for hashing over a typical convolutional neural network are as follows: 
\begin{itemize}[leftmargin=*]
\item \textbf{Training of very deep networks}: The representational power of deep networks should ideally increase with increased depth. It is empirically observed that in deep feed-forward nets beyond a certain depth, adding additional layers results in higher training and validation error (despite using batch normalization)~\cite{ResNet}. Residual networks seamlessly solves this \textit{via} adding short cut connections that are summed with the output of the convolutional blocks. 

\item \textbf{Ease of Optimization}: A major issue to training deep architectures is the problem of vanishing gradients during training (this is in part mitigated with the introduction of rectified linear units (ReLU), input batch normalisation and layer normalisation). Residual connections offer additional support \textit{via} a no-resistance path for the flow of gradients along the shortcut connections to reach the shallow learning layers. 
\end{itemize}

\subsection{Supervised Retrieval Loss Function}


In order to learn feature embeddings tailored for retrieval and specifically for the scenario at hand where the pairwise similarity matrix $\mathcal{S}$ should be preserved, we propose our supervised retrieval loss drawing inspiration from the neighbourhood component analysis~\cite{NCA}. To encourage the learnt embedding to be binary in nature, we squash the output of the residual layers to be within $\left [-1, 1 \right ]$ by passing it through a hyperbolic tangent (tanh) activation function. The final binary hash codes $(\mathbf{b}_{i})$ are generated by quantizing the output of the tanh activation function (say, $\mathbf{h}_{i}$) as follows: $\mathbf{b}_{i} = \text{sgn}\left ( \mathbf{h}_{i} \right )$. Given $N$ instances and the corresponding similarity matrix is defined as $\mathcal{S} = \left \{ s_{ij} \right \}_{i,j=1}^{N} \in \left \{ 0,1 \right \}^{N \times N}$,
the proposed supervised retrieval loss is formulated as:
\begin{equation}
J_{S} = 1 - \frac{1}{N}\sum_{i,j = 1}^{N}p_{ij}s_{ij} 
\label{eq:lossS}
\end{equation}

\noindent
where $p_{ij}$ is the probability that any two instances ($i$ and $j$) can be potential neighbours. Inspired by kNN classification, where the decision of an unseen test sample is determined by the semantic context of its local neighbourhood in the embedding space, we define $p_{ij}$ as a softmax function of the hamming distance (indicated as $\oplus$) between the hash codes of two instances and is derived as:
\begin{equation}
p_{ij} = \frac{e^{-\left ( \mathbf{b}_{i} \oplus \mathbf{b}_{j} \right )}}{\sum_{l\neq i}e^{-\left ( \mathbf{b}_{i} \oplus \mathbf{b}_{l} \right )}} \text{ where } \mathbf{b}_{\left ( \cdot \right )} = \text{sgn}\left ( \mathbf{h}_{\left ( \cdot \right )} \right )
\label{eq:pij}
\end{equation}

As gradient based optimisation of  $J_{s}$ in a binary embedding space is infeasible due to its non-differentiable nature, we use a continuous domain relaxation and substitute non-quantized embeddings $\mathbf{h}_{\left ( \cdot \right )}$ in place of hash code $\mathbf{b}_{\left ( \cdot \right )}$ and Euclidean distance as as surrogate of Hamming distance between binary codes. This is derived as: $p_{ij} = e^{-\left \| \mathbf{h}_{i} - \mathbf{h}_{j} \right \|^{2}} / \sum_{i \neq l}e^{-\left \| \mathbf{h}_{i} - \mathbf{h}_{l} \right \|^{2}}$. It must be noted that such an continuous relaxation could potentially result in uncontrollable quantization error and large approximation errors in distance estimation. With continuous relaxation, Eq.~\eqref{eq:lossS} is now differentiable and continuous thus suited for backpropagation of gradients during training.

\subsection{Hashing related Loss Functions and Regularization}

Generation of high quality hash codes requires us to control this quantization error and bridge the gap between the Hamming distance and its continuous surrogate. In this paper, we jointly optimise for $J_{s}$ and improve hash code generation by imposing additional loss functions as follows:

\textbf{Quantization Loss}: In the seminal work on iterative quantization (ITQ) for hashing~\cite{itq}, Gong and Lazebnik introduced the notion of quantization error $J_{Q-\text{ITQ}}$ as $J_{Q-\text{ITQ}} = \left \| \mathbf{h}_{i} - \text{sgn}\left ( \mathbf{h}_{i} \right ) \right \|_{2}$.
Optimising for $J_{Q-\text{ITQ}}$ required a computation intensive alternating optimisation procedure and is not compatible with back propagation which is used to train deep neural nets (due to non-differentiable sgn function within the formulation). Towards this end, we use a modified point-wise quantization loss function proposed by Zhu \textit{et al.} sans the sgn function as $J_{Q -\text{Zhu}} = \left \| \left | \mathbf{h}_{i} \right | - \mathbf{1} \right \|_{1}$~\cite{zhu2016}. They establish that $J_{Q -\text{Zhu}}$ is an upper bound over $J_{Q-\text{ITQ}}$, therefore can be deemed as a reasonable loss function to control quantization error. For ease of back-propagation, we propose to use a differentiable smooth surrogate to $L_{1}$ norm $\left | \left ( \cdot \right ) \right |_{1} \approx \text{log cosh}\left ( \cdot \right )$ and derived the proposed quantization loss function as:$J_{Q} = \sum_{i=1}^{N} \left ( \text{log cosh}\left (\left | \mathbf{h}_{i} \right | - \mathbf{1} \right ) \right )$.
With the incorporation of the quantization loss, we hypothesise that the final binarization step would incur significantly less quantization error and the loss of retrieval quality (also empirically validated in Section~\ref{sec:Results}).

\textbf{Bit Balance Loss}: In addition to $J_{Q}$, we introduce an additional bit balance loss $J_{B}$ to maximise the entropy of the learnt hash codes and in effect create balanced hash codes. Here, $J_{B}$ is derived as:
$J_{B} = -\frac{1}{2N}\text{tr}\left ( \mathbf{H}\mathbf{H}^{T} \right )$. This loss aims at encouraging maximal information storage within each hash bit.

\textbf{Regularisation}: Inspired by ITQ~\cite{itq}, we also introduce a relaxed orthogonality regularisation constraint $R_{O}$ on the convolutional weights (say, $\mathbf{W}_{h}$) connecting the output of the final residual block of the network to the hashing block. This weakly enforces that the generated codes are not correlated and each of the hash bits are independent. Here, $R_{O}$ is formulated as:
$R_{O} = \frac{1}{2} \left \| \mathbf{W}_{h}\mathbf{W}_{h}^{T} - \mathbf{I} \right \|_{F}^{2}$.
In additon to $R_{O}$, we also impose weight decay regularization  $R_{W}$ to control the scale of learnt weights and biases.


\subsection{Model Learning}

In this section, we detail on the training procedure for the proposed DRH network with respect to the supervised retrieval and hashing related loss functions. We learn a single-stage end-to-end deep network to generate hash codes directly given an input image. We formulate the optimisation problem to learn the parameters of our network (say, $\Theta : \left \{ \mathbf{W}^{\left ( \cdot \right )}, b^{\left ( \cdot \right )} \right \}$):

\begin{equation}
\underset{\Theta : \left \{ \mathbf{W}^{\left ( \cdot \right )}, b^{\left ( \cdot \right )} \right \}}{\text{arg}\text{min}} J = J_{S} + \underbrace{ \lambda_{q}J_{Q} + \lambda_{b}J_{B} }_{\text{Hashing Losses}} + \underbrace{ \lambda_{o}R_{O} + \lambda_{w}R_{W}}_{\text{Regularisation}}
\label{eq:JointLoss}
\end{equation}
\noindent
where $\lambda_{q}$, $\lambda_{b}$, $\lambda_{o}$ and $\lambda_{w}$ are four parameters to balance the effect of different contributing terms. To solve this optimisation problem, we employ stochastic gradient descent to learn optimal network parameters. Differentiating $J$ with respect to $\Theta$ and using chain rule, we derive:
$\frac{\partial J}{\partial \Theta } = \frac{\partial J}{\partial \mathbf{H}}\frac{\partial \mathbf{H}}{\partial \Theta} = \frac{1}{N}\sum_{i =1}^{N}\frac{\partial J}{\partial \mathbf{h}_{i}}\frac{\partial \mathbf{h}_{i}}{\partial \Theta}$
The second term $\partial \mathbf{h}_{i}/\partial \Theta$ is computed through gradient back-propagation. The first term ($\partial J / \partial \mathbf{h}_{i}$) is the gradient of the composite loss function $J$ with respect to the output hash codes of the DRH network.

%

We differentiate the continuous relaxation of the supervised retrieval loss function with respect to the hash code of a single example ($ \mathbf{h}_{i}$) as follows~\cite{NCA}:
\begin{dmath}
\frac{\partial J_{S}}{\partial \mathbf{h}_{i}} = 2\left ( \sum_{l:s_{li} > 0} p_{li}d_{li} - \sum_{l \neq i}\left ( \sum_{q:s_{lq} > 0} p_{lq} \right ) p_{li}d_{li} \right ) - 2\left ( \sum_{j:s_{ij} > 0} p_{ij}d_{ij} - \sum_{j:s_{ij} > 0} p_{ij}\left ( \sum_{z \neq i} p_{iz}d_{iz} \right ) \right )
\label{eq:derJS}
\end{dmath}
\noindent
where $d_{ij} = \mathbf{h}_{i} - \mathbf{h}_{j}$. The derivatives of hashing related loss functions ($J_{Q}$ and $J_{B}$) are derived as:
$\frac{\partial J_{Q}}{\partial \mathbf{h}_{i}} = \text{tanh}\left ( \left | \mathbf{h}_{i} \right | - \mathbf{1} \right )\text{sgn}\left ( \mathbf{h}_{i} \right )$ and $\frac{\partial J_{B}}{\partial \mathbf{h}_{i}} = -\mathbf{h}_{i}$
The regularisation function $R_{O}$ acts on the convolutional weights corresponding to the hash layer ($\mathbf{W}_{h}$) and its derivative with respect to $\mathbf{W}_{h}$ is derived as follows:
$\frac{\partial R_{O}}{\partial \mathbf{W}_{h}} = \mathbf{W}_{h}\left ( \mathbf{W}_{h}\mathbf{W}_{h}^{T} - \mathbf{I} \right )$. 
\noindent

Having computed the gradients of the individual components of the loss function with respect to the parameters of DRH, we apply gradient-based learning rule to update $\Theta$. We use mini-batch stochastic gradient descent (SGD) with momentum. SGD incurs limited memory requirements and reduces the variance of parameter updates. The addition of the momentum term $\gamma$ leads to stable convergence. The update rule for the weights of the hash layer is derived as:

\begin{equation}
\mathbf{W}_{h}^{t} = \mathbf{W}_{h}^{t-1} - \nu^{t} \text{ where }
\nu^{t} = \gamma\nu^{t-1} + \eta \left ( \frac{\partial J}{\partial \mathbf{W}_{h}^{t-1} } + \lambda_{o} \frac{\partial R_{O}}{\partial \mathbf{W}_{h}^{t-1}} +\lambda_{w}\frac{\partial R_{W}}{\partial \mathbf{W}_{h}^{t-1}} \right ) 
\label{eq:updateWh}
\end{equation}
\noindent
The convolutional weights and biases of the other layers are updated similarly. It must be noted that the learning rate $\eta$ in Eq~\ref{eq:updateWh} is an important hyper-parameter. For faster learning, we initialise it the largest learning rate that stably decreases the objective function (typically, at $10^{-2}$ or $10^{-3}$). Upon convergence at a particular setting of $\eta$, we scale the learning rate multiplicatively by a factor of $0.1$ and resume training.This is repeated until convergence or reaching the maximum number of epochs. 
%


\section{Experiments and Observations}
\label{sec:experiment}


\noindent
\textbf{Database}:
We conducted empirical evaluations on the publicly available Indiana University Chest X-rays (CXR) dataset
%
archived from their hospital's picture archival systems~\cite{IndianaCXR}. The fully-anonymized dataset is publicly available through the OpenI image collection system~\cite{openI}. For this paper, we use a subset of 2,599 frontal view CXR images that have matched radiology reports available for different patients. Following the label generation strategy published in~\cite{shin2016cvpr} for this dataset, we extracted nine most frequently occurring unique patterns of Medical Subject Headings (MeSH) terms related to cardiopulmonary diseases from these expert-annotated radiology report~\cite{mesh}. 
These include normal, opacity, calcified granuloma, calcinosis, cardiomegaly, granulomatous disease, lung hyperdistention, lung hypoinflation and nodule. The dataset was divided into non-overlapping subsets for training (80\%) and testing (20\%) with patient-level splits. The semantic similarity matrix $\mathcal{S}$ is contructed using the MeSH terms \textit{i.e.} a pair of images are considered similar if they share atleast one MeSH term.

\noindent
\textbf{Comparative Methods and Baselines}:
We evaluate and compare the retrieval performance of the proposed DRH network with nine state-of-the art methods including five unsupervised shallow-learning methods: LSH~\cite{lsh}, ITQ~\cite{itq}, CBE~\cite{cbe}; two supervised shallow-learning methods: KSH~\cite{ksh} and MHF~\cite{mhf} and two deep learning based methods: AlexNet - KSH (A - KSH)~\cite{alexNet} and VGGF - KSH (V - KSH)~\cite{VGGF}. To justify the proposed formulation, we include simplified four variants of the proposed DRH network as baselines: DPH (Deep Plain Net Hashing) by removing the residual connections, DRHNQ (Deep Residual Hashing without Quantization) by removing the hashing related losses and generating binary codes only through tanh activation, DRN - KSH by training a deep residual network with only the supervised retrieval loss and quantizing through KSH post training and DRH - NB which is a variant of DRH where continuous embeddings are used sans quantization, which may act as an upper bound on performance. We used the standard metrics for evaluating retrieval quality as proposed by Lai \textit{et al.}~\cite{lai2015}: Mean Average Precision (MAP) and Precision - Recall Curves varying the code size(16, 32, 48 and 64 bits). For fair comparison, all the methods were trained and tested on identical data folds. The retrieval performance of methods involving residual learning and  baselines is evaluated for two variants varying the number of layers: $\left ( \cdot \right ) - 18$ and $\left ( \cdot \right ) - 34$.

For the shallow learning methods, we represent each image as a 512 dimensional GIST vector~\cite{gist}. For the DRH and associated baselines, the input image is resized to $224 \times 224$ and normalized to a dynamic range of 0-1 using the pre-processing steps discussed in~\cite{shin2016cvpr}. For A-KSH and V-KSH, the image normalization routines were identical to that reported in the original works~\cite{alexNet}~\cite{VGGF}. We implement all our deep learning networks ( including DRH) on the open-source MatConvNet framework~\cite{MatConvNet}. The hyper-parameters $\lambda_{q}$, $\lambda_{b}$ and $\lambda_{0}$ were set at 0.05, 0.025 and 0.01 empirically. The momentum term $\gamma$ was set at 0.9, the initial learning rate $\eta$ at $10^{-2}$ and batchsize at 128. The training data was augmented on-the-fly extensively through jittering, rotation and intensity augmentation by matching histograms between images sharing similar co-morbidities. All the comparative deep learning methods were also trained with similar augmentation. Furthermore, for A - KSH and V - KSH variants, we pre-initialized the network parameters from the pre-trained models by removing the final probability layer~\cite{alexNet}~\cite{VGGF}. These network learnt a $4096$-dimensional embedding by fine-tuning it with cross-entropy loss. The hashing was performed explicitly through KSH upon convergence of the network.

\noindent
\textbf{Results}:
\label{sec:Results}
The results of the MAP of the Hamming ranking for varying code sizes of all the comparative methods are listed in Table~\ref{wrap-tab:1}.  We report the precision-recall curves for the comparative methods at a code size of 64 bits in Fig.~\ref{fig:PRComp}.
To justify the proposed formulation for DRH, several variants of DRH (namely, DRN - KSH, DPH, DRH - NQ and DRH - NB) were investigated and compare their retrieval results are tabulated in Table~\ref{wrap-tab:2}. In addition to MAP, we also report the retrieval precision withing Hamming radius of 2 (P @ H2).  The associated precision-recall curves are shown in Fig.~\ref{fig:PRBL}.

\vspace{8pt}
\begin{minipage}[!t]{\textwidth}
\hspace{-0.7cm}
\begin{minipage}{0.45\textwidth}
\centering
\includegraphics[width=\textwidth]{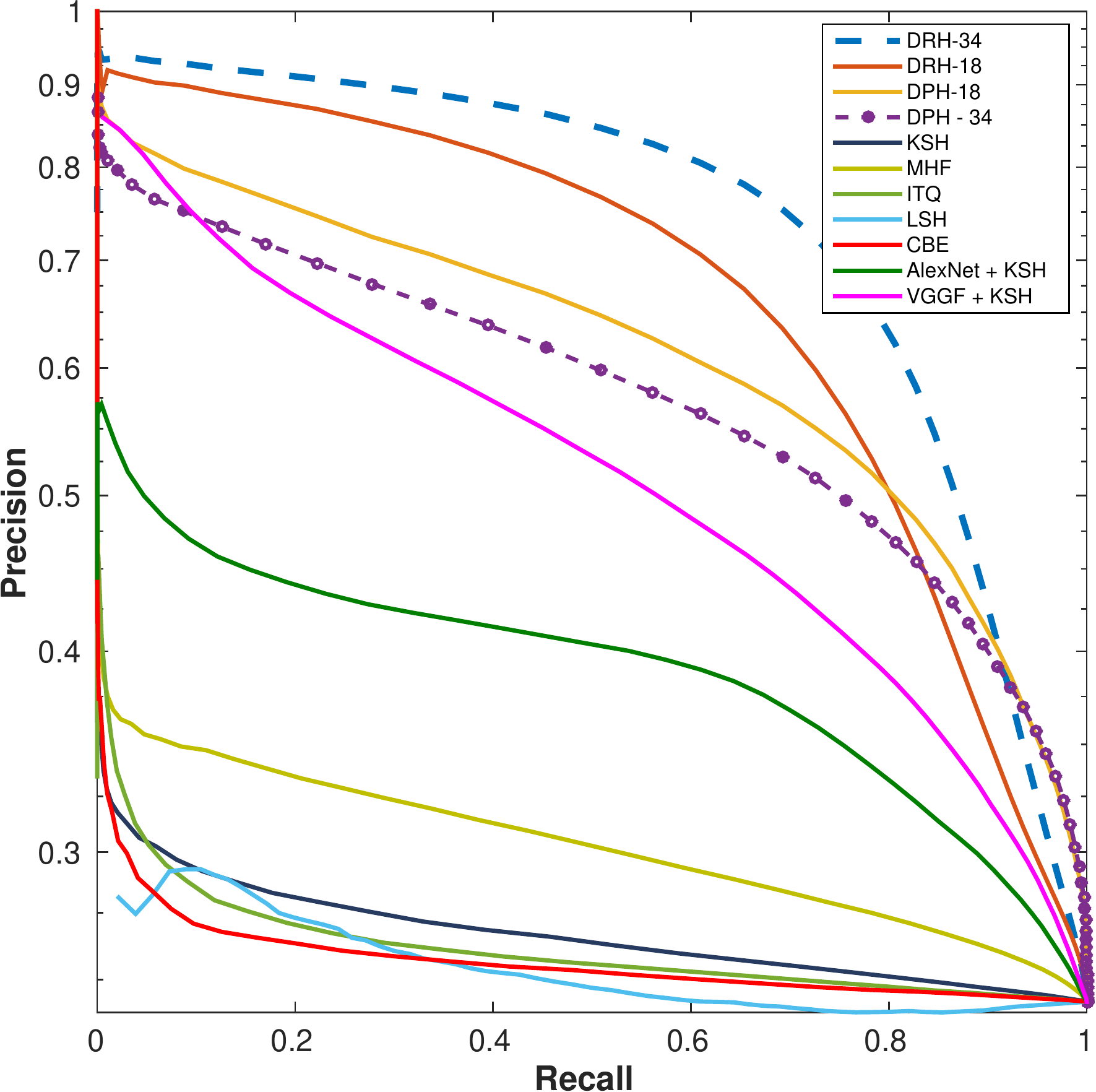}
\captionof{figure}{PR Curves at code size of 64 bits for the comparative methods.}
\label{fig:PRComp}
\end{minipage}
\hfill
\begin{minipage}{0.45\textwidth}
\centering
\hspace{-1.7cm}
\includegraphics[width=\textwidth]{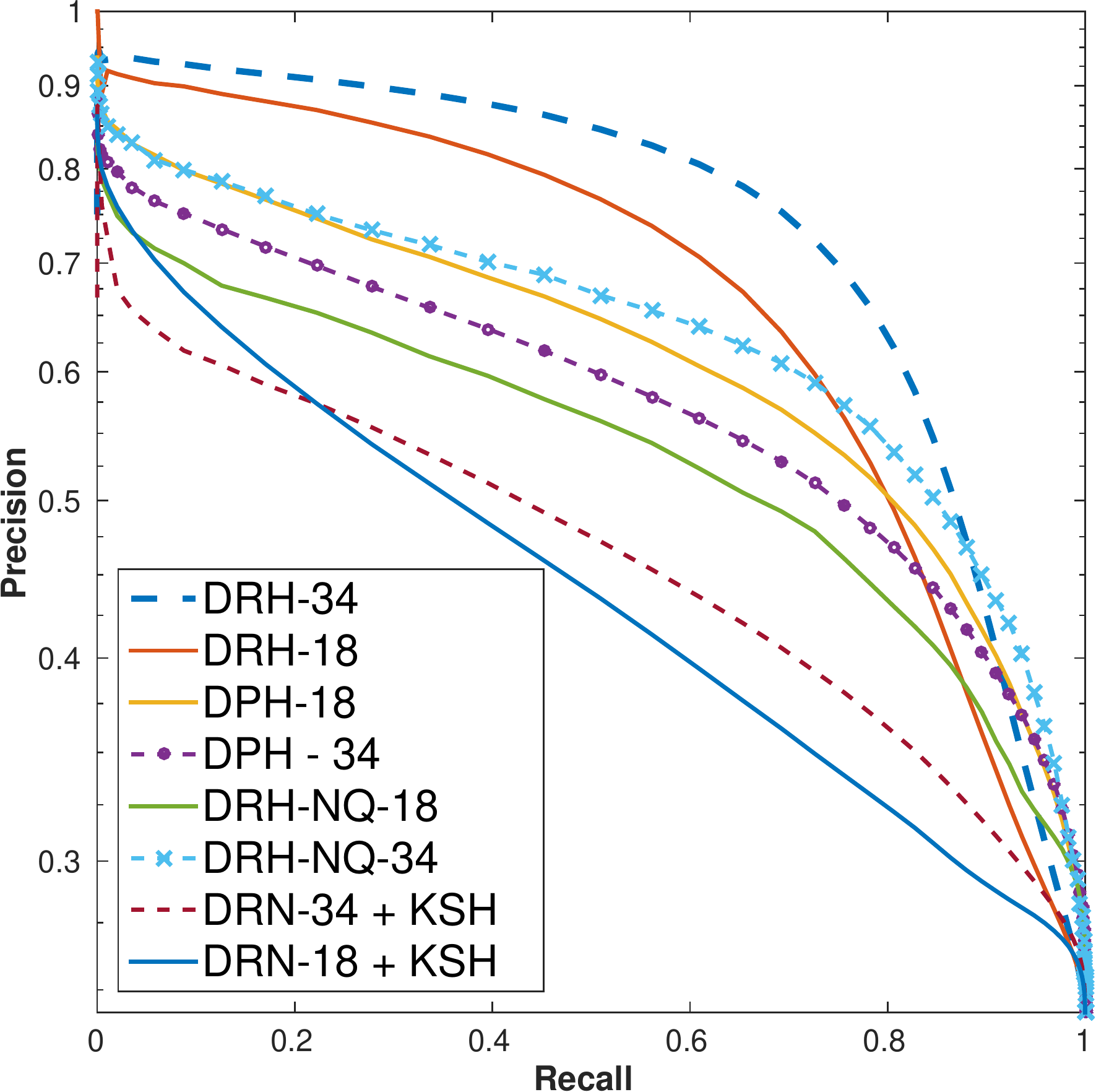}
\captionof{figure}{PR Curves at code size of 64 bits for the baseline variants of DRH.}
\label{fig:PRBL}
\end{minipage}
\end{minipage}

\section{Discussion}
\label{sec:Discussion}

Within this section, we present our discussion answering the questions posed in Section~\ref{sec:intro}, w.r.t. to the results and observations we reported in Section~\ref{sec:Results}.

\noindent
\textbf{Trainability}: The introduction of residual connections offers short-cut connections which act as zero-resistance paths for gradient flow thus effectively mitigating vanishing of gradients as network depth increases. This is strongly substantiated by comparing the performance of DRH - 34 to DRH - 18 \textit{vs.} the plain net variants of the same depth DPH - 34 to DPH - 18. There is a strong improvement in MAP with increasing depth for DRH of about 9.3\%. On the other hand, we observe a degradation of 2.2\% MAP performance on increasing layer depth in DPH. The performance of DRH-18 is fractionally better than DPH - 18 indicating that DRH exhibits better generalizability and the degradation problem is addressed well as we have significant MAP gains from increased depth.
 With the introduction of batch normalisation and residual connections, we ensure that the signals during forward pass have non-zero variances and that the back propagated gradients exhibit healthy norms. Therefore, neither forward nor backward signals vanish within the network. This is substantiated by the differences in MAP observed in Table~\ref{wrap-tab:1} between methods using BN (DRH, DPH and V-KSH) in comparison to A-KSH which does not use BN.

\noindent
\textbf{Representability}: Ideally, the latent embeddings in the Hamming space should be such that similar samples are mapped closer while simultaneously mapping dissimilar samples further apart. We plot the t-Stochastic Neighbourhood Embeddings (t-SNE)~\cite{tSNE} of the hash codes for four comparative methods ( GIST - ITQ, VGGF - KSH, DPH - 18 and DRH - 34) in Fig.~\ref{fig:netTSNE} to visually assess the quality of the hash codes generated. Visually, we observe that hand-crafted GIST features with unsupervised hashing method ITQ fail to sufficiently induce semantic separability. In comparison, though VGGF-KSH improves significantly owing to network fine-tuning, better embedding results from DRH - 34 (DPH-18 is highly comparable to DRH-34). Additionally, the significant differences in MAP reported in Table~\ref{wrap-tab:1} between these methods substantiates our hypothesis that in scenarios of limited training data it is better to train smaller models from scratch over finetuning to avoid overfitting (DRH - 34 has 0.183M in comparison to VGGF with 138M parameters). Also the significant domain shift between natural images (ImageNet - VGGF) and CXR poses a significant challenge for generalizability of networks finetuned from pre-trained nets.

\begin{wraptable}{l}{5.6cm}
\vspace{-12pt}
\resizebox{5.5cm}{!}{\begin{tabular}{c|c|c|c|c}
\multirow{2}{*}{Method} & \multicolumn{4}{c}{MAP} \\ \cline{2-5}
& 16 bits & 32 bits & 48 bits & 64 bits \\ \hline
LSH & 22.77 & 23.93 & 23.99 & 24.85 \\
ITQ & 25.06 & 25.19 & 25.87 & \textbf{26.23} \\
CBE & 13.21 & 26.00 & 25.32 & 25.69 \\\hline
MHF & 23.62 & 27.02 & 30.78 & \textbf{36.75} \\
KSH & 26.46 & 32.49 & 32.01 & 30.42 \\ \hline
A - KSH & 35.95 & 37.28 & 36.64 & 39.31 \\
V - KSH & 47.92 & 50.64 & \textbf{53.62} & 52.61 \\ \hline
DPH - 18 & 48.78 & 52.13 & 54.01 & \textbf{66.59} \\
DPH - 34 & 46.64 & 44.43 & 51.39 & 64.38 \\ \hline
\textbf{DRH - 18} & 50.93 & 57.46 & 62.76 & 67.44 \\
\textbf{DRH - 34} & 56.79 & 65.80 & 75.81 & \textbf{76.72} \\ \hline
\end{tabular}}
\caption{MAP of Hamming ranking w.r.t. varying code sizes for comparative methods.}\label{wrap-tab:1}
\resizebox{5.5cm}{!}{\begin{tabular}{c|c|c|c|c}
\multirow{2}{*}{Method} & \multicolumn{2}{c|}{MAP} & \multicolumn{2}{c}{P @ H2} \\ \cline{2-5}
& 18-L & 34-L & 18-L & 34-L \\ \hline

DRN - KSH & 54.60 & 62.50 & 83.58 & 87.50 \\
DPH & 66.59 & 64.38 & 91.50 & 93.10 \\
DRH - NQ & 62.69 & 66.23 & 82.37 & 89.32 \\
DRH - NB & 69.21 & 77.45 & 95.81 & 95.64 \\
\textbf{DRH} & 67.44 & \textbf{76.72} & 95.56 & 94.59 \\ \hline
\end{tabular}}
\caption{MAP and P @ H2 of the Hamming ranking w.r.t. varying network depths for baseline variants of DRH at a fixed code size of 64 bits.}\label{wrap-tab:2}
\vspace{-24pt}
\end{wraptable}

\noindent
\textbf{Compactness}: Hashing aims at generating compact representations preserving the semantic relevance to the maximal extent. Varying the code sizes, we observe from Table~\ref{wrap-tab:1} that the MAP performance of majority of the supervised hashing methods improves significantly. In particular for DRH - 34, we observe that the improvement in the performance from 48 bits to 64 bits is only fractional. The performance of DRH - 34 at 32 bits is highly comparable to DRH - 18 at 64 bits. This testifies that with increasing layer depth DRH learns more compact binary embeddings such that shorter codes can already result in good retrieval quality.

\begin{wrapfigure}{r}{0.5\textwidth}
\vspace{-18pt}
\includegraphics[width =0.5\textwidth]{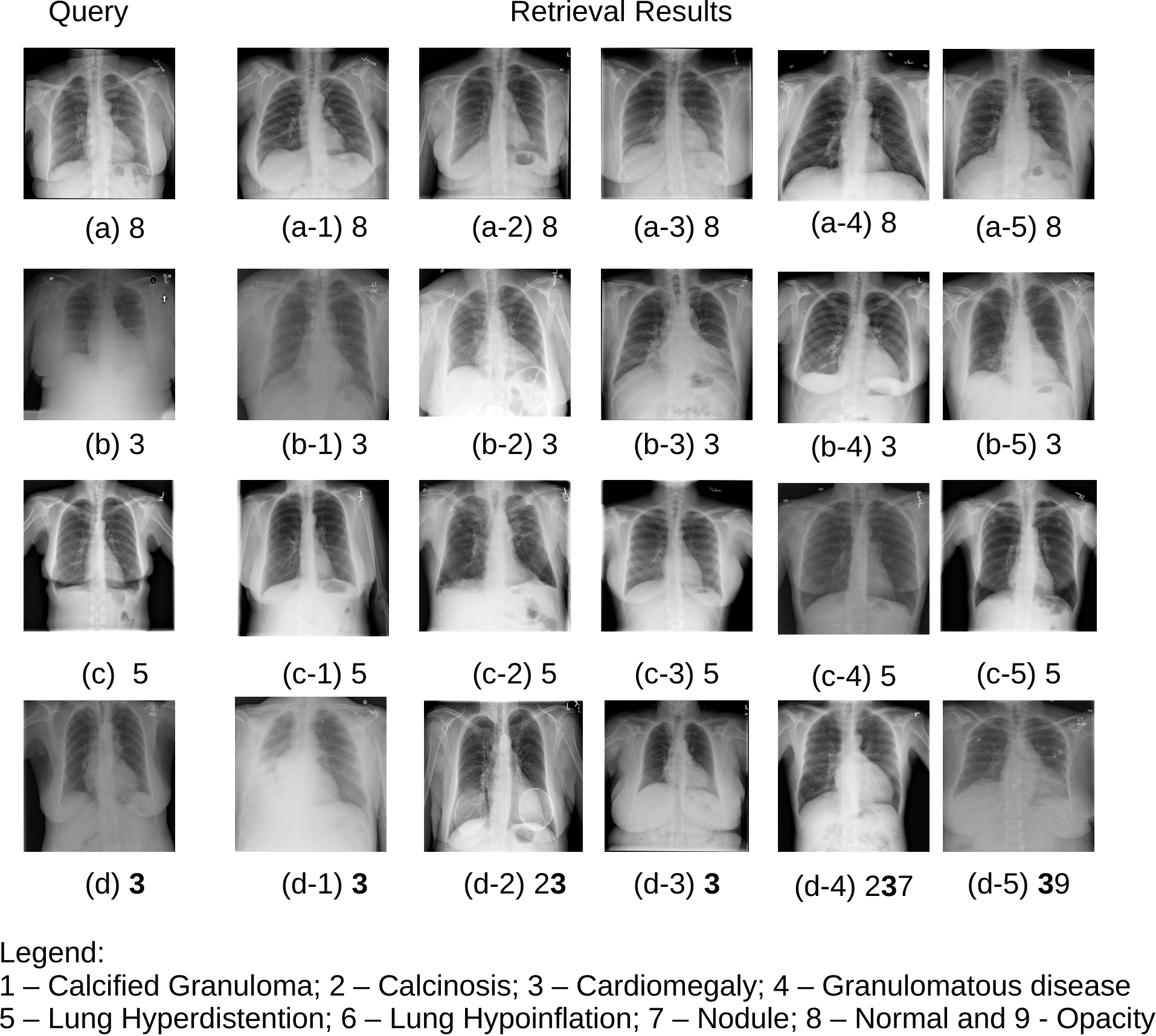}
\caption{Retrieval results for DRH-34.}
\label{fig:netResults}
\vspace{-24pt}
\end{wrapfigure}

\noindent
\textbf{Semantic Similarity Preservation}:   Visually assessing the t-SNE representation of  GIST - ITQ (Fig.~\ref{fig:netTSNE}(a)) we can observe that it fails to sufficiently represent the underlying semantic relevance within the CXR images in the latent hamming space, which retestifies the  concerns over hand-crafted features that were raised in Section~\ref{sec:motivation}. VGGF - KSH (Fig.~\ref{fig:netTSNE}(b)) improves over GIST - ITQ substantially, however it fails to induce sufficient class-separability. Despite KSH considering pair-wise relationships while learning to hash, the feature representation generated by fine-tuned VGG-F is limited in representability as the cross-entropy loss is evaluated point-wise. Finally, the tSNE embedding of DRH - 34  shown in Fig.~\ref{fig:netTSNE} visually reaffirms that semantic relevance remains preserved upon embedding and the method generates clusters well separated within the hamming space. The high degree of variance associated with the tSNE embedding of normal class (red in color) is conformal with the high population variability expected within that class.
Fig.~\ref{fig:netResults} demonstrates the first five retrieval results sorted according to their Hamming rank for four randomly selected CXR images from the testing set. In particular, for Case (d), where we observe that the top neighbours (d 1-5) share atleast one co-occurring pathology. For cases (a), (b) and (c), all the top five retrieved neighbours share the same class. 

\noindent
\textbf{Joint Optimisation}: The main contribution of the work hinges on the hypothesis that performing an end-to-end learning of hash codes is better than a two stage learning process. Comparative validations against the two-stage deep learning methods (A - KSH, V - KSH and baseline variant DRN - KSH) strongly support this hypothesis. In particular, we observe over 14.2\% improvement in MAP comparing DRN - KSH (34 - L) to DRH - 34. This difference in performance may be owed to a crucial disadvantage of DRN - KSH that the generated feature representation is not optimally compatible to binararization. We can also observe that, DRH - 18 and DRH - 34 incur very small average MAP decrease fo 1.8\% and 0.7\% when binarizing hash codes against non-binarized continuous embeddings in DRH - B- 18 and DRH - B - 34 respectively. In contrast, DRH - NQ suffers from very large MAP decreases of 6.6\% and 10.8\% in comparison to DRH - B. These observations validate the need for the proposed quantization loss as it leads to nearly lossless binarization.

%

\section{Conclusions and Open Questions}

In this paper, we have presented a novel deep learning based hashing approach leveraging upon residual learning, termed as Deep Residual Hashing (DRH). DRH integrates representation learning and hash coding into a joint optimisation framework with dedicated losses for improving retrieval performance and hashing related losses to control the quantization error and improve the hash code quality. Our approach demonstrated very promising results on a challenging chest x ray dataset with co-occurring morbidities. Taking insights from this pilot study on retrieval of CXR images with cardiopulmonary diseases, we believe gives rise to the following open questions for further discussion: How deep is deep enough? How does DRH extend to include an additional anatomical view ( like the dorsal view for CXR) improve retrieval performance? Does DRH generalize to unseen disease manifestations?; and Can we visualize what DRH learns? In conclusion, we believe that our paper strongly supports our initial premise of using DRH for retrieval but also opens up questions for future discussions.

%
%
%


\end{document}